\documentclass[10pt,twocolumn,letterpaper]{article}

\usepackage{cvpr}              

\usepackage{graphicx}
\usepackage{amsmath}
\usepackage{amssymb}
\usepackage{booktabs}

\usepackage{gensymb}
\usepackage{dsfont}
\usepackage{wrapfig,lipsum,booktabs}
\usepackage{caption}
\usepackage{epsfig}
\usepackage{graphicx}
\usepackage{amsmath}
\usepackage{amsthm}
\usepackage{amssymb}
\usepackage{dsfont}
\usepackage{color}

\usepackage{afterpage}
\usepackage{float}
\usepackage[T1]{fontenc}
\usepackage{cite}
\usepackage{nopageno}
\usepackage{xcolor}

\usepackage[pagebackref,breaklinks,colorlinks]{hyperref}

\usepackage[capitalize]{cleveref}
\crefname{section}{Sec.}{Secs.}
\Crefname{section}{Section}{Sections}
\Crefname{table}{Table}{Tables}
\crefname{table}{Tab.}{Tabs.}

\def\cvprPaperID{4444} 
\def\confName{CVPR}
\def\confYear{2023}

\begin{document}

\title{Normal-guided Garment UV Prediction for Human Re-texturing}
\vspace{-5mm}
\author{
Yasamin Jafarian$^\dagger$
\hspace{5mm}Tuanfeng Y. Wang$^\sharp$
\hspace{5mm}Duygu Ceylan$^\sharp$
\hspace{5mm}Jimei Yang$^\sharp$
\vspace{1mm}
\\
\hspace{5mm}Nathan Carr$^\sharp$
\hspace{5mm}Yi Zhou$^\sharp$
\hspace{5mm}Hyun Soo Park$^\dagger$
\vspace{3mm}
\\
\hspace{-0mm}$^\dagger$University of Minnesota
\hspace{10mm}
$^\sharp$Adobe Research \\
}


\twocolumn[{
\renewcommand\twocolumn[1][]{#1}
\maketitle
\begin{center}
    \centering
    \vspace{-5mm}
    \includegraphics[width=1\textwidth]{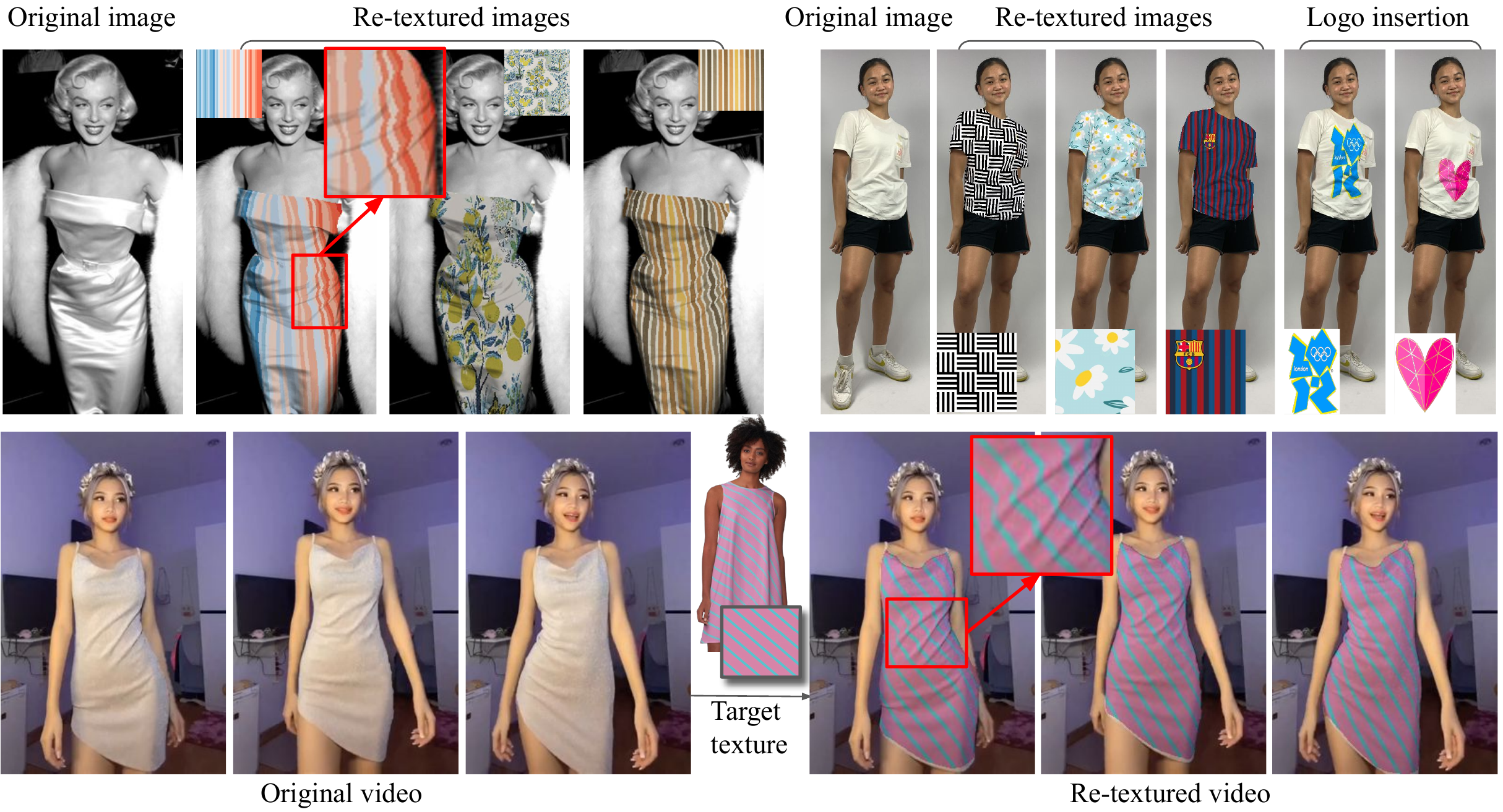}
    \vspace{-7mm}
    \captionof{figure}{\textbf{Geomety aware texture editing from images and vidoes.}
    This paper presents a novel approach to predict a geometry aware texture (UV) map of a garment from an image (first row) and video (second row). The predicted UV map preserves isometry between texture space and 3D surface by leveraging 3D surface normals. Further, we ensure temporal consistency in the predicted UV map across frames in a video, resulting in physically plausible human appearance editing. }
    \label{fig:teaser}
\end{center}}]

\begin{abstract}
\vspace{-5mm}
Clothes undergo complex geometric deformations, which lead to appearance changes. To edit human videos in a physically plausible way, a texture map must take into account not only 
the garment transformation induced by the body movements and clothes fitting,
but also its 3D fine-grained surface geometry. This poses, however, a new challenge of 3D reconstruction of dynamic clothes from an image or a video. In this paper, we show that it is possible to edit dressed human images and videos without 3D reconstruction. We estimate a geometry aware texture map between the garment region in an image and the texture space, a.k.a, UV map. Our UV map is designed to preserve isometry with respect to the underlying 3D surface by making use of the 3D surface normals predicted from the image. Our approach captures the underlying geometry of the garment in a self-supervised way, requiring no ground truth annotation of UV maps and can be readily extended to predict temporally coherent UV maps. We demonstrate that our method outperforms the state-of-the-art human UV map estimation approaches on both real and synthetic data.
\end{abstract}
\vspace{-10mm}

\section{Introduction}


While browsing online clothing shops, have you ever wondered how the appearance of a dress of interest would look on you as if you were in a fitting room {given your dress with a similar shape}?
A key technology to enable generating such visual experiences is \textit{photorealistic re-texturing}---editing the texture of clothes in response to the subject's movement in the presented images or videos in a geometrically and temporally coherent way.
Over the past few years, there has been a significant advancement in the image and video editing technologies \cite{ling2021editgan, wu2020cascade, Navigan_CVPR_2021, shen2020interpreting, bau2019semantic, chen2020deepfacedrawing, alharbi2020disentangled, collins2020editing, kim2021stylemapgan, lee2020maskgan, hou2020guidedstyle, he2019attgan, mir20pix2surf, bhatnagar2019mgn, Lazova2019360DegreeTO, grigorev2019coordinate}, such as inserting advertising logos on videos of moving cars or applying face makeup on social media. 
However, such editing approaches designed for rigid or semi-rigid surfaces are not suitable for garments that undergo complex secondary motion with respect to the underlying body. For example, the fine wrinkles of the dress in Figure \ref{fig:teaser} result in complex warps in texture over time. 
In this paper, we present a new method to edit the appearance of a garment in a given image or video by taking into account its fine-grained geometric deformation. 

Previous works address photorealistic texture editing in two ways. (1) 3D reconstruction and rendering: 
these approaches can achieve high-fidelity texture editing given highly accurate 3D geometry. On other side of the coin, their performance is dictated by the quality of the 3D reconstruction. While the 3D geometry of the garment can be learned from paired human appearance data, e.g., human modeling repositories with 3D meshes and renderings~\cite{RP:2020}, due to the scarcity of such data, it often cannot generalize well on unseen real images and videos.
(2) Direct texture mapping: by estimating dense UV map, these methods can bypass the procedure of 3D reconstruction~\cite{guler2018densepose, Neverova2020ContinuousSurfaceEmbeddings, Neverova2021UniversalCanonicalMaps, ianina2022bodymap, Xie_2022_CVPR}. However, they usually lack of geometry details and only capture the underlying human body, thus, not applicable for editing garments. Moreover, when applied to videos, visual artifacts of editing become more salient since they are not aware of underlying deformation of the garment's 3D geometry \cite{kasten2021layered,ye2022sprites
}.
We design our method to enjoy the advantages of both two approaches: preserving realistic details in UV mapping while circumventing 3D reconstruction. Our key insight is that the fundamental geometric property of isometry can be imposed into UV map estimation via the 3D surface normals predicted from an image. We formulate a geometric relationship between the UV map and surface normals in the form of a set of partial differential equations. 

Our method takes as input an image or video, its surface normal prediction, and dense optical flow (for video), and outputs the geometry aware UV map estimate. 
The UV map is modeled by a multi-layer perceptron that can predict UV coordinates given a pixel location in an image. We note that the UV map is defined up to the choice of a reference coordinate frame. To disambiguate this, we condition the neural network with a pre-defined proxy UV map (e.g., DensePose~\cite{guler2018densepose}). 
We use the isometry constraints as a loss to optimize the UV map.
Further, for a video, we leverage the per-frame image feature to correlate the UV coordinates of the pixels across time using optical flow.

Our contributions can be concluded in three aspects:
(1) a novel formulation that captures the geometric relationship between the 3D surface normals and the UV map by the  isometry contraint, which eliminates the requirement of 3D reconstruction and ground truth UV map;
(2) a neural network design that learns to predict temporally coherent UV map for the frames by correlating per-frame image features; 
(3) stronger performance 
compared to existing 
re-texturing methods and compelling results on a wide range of real-world imagery.


\section{Related Work}
\label{sec:relatedwork}

Our work lies at the intersection of human UV map prediction from images and neural UV map optimization.

\subsection{Human Dense UV Map Estimation}

A seminal work of DensePose~\cite{guler2018densepose} learns to predict a UV map of humans presented in an image, which opens a new opportunity to edit the appearance of a person without 3D reconstruction~\cite{albahar2021pose, Wang2021dance}. 
A series of subsequent works \cite{guler2018densepose,Neverova2020ContinuousSurfaceEmbeddings, Neverova2021UniversalCanonicalMaps, zeng20203d, Yan_2021_ICCV, NEURIPS2019_53fde96f, Neverova2019SlimDT, Yu2021SemisupervisedDK, Kulkarni0FT20, kulkarni2020articulation} bring out a number of applications for human tracking. 
However, due to their representation specific to the body surface, they exhibit fundamental limitations in expressing highly deformable loose clothing such as skirts and dresses.

To address this challenge, recent approaches leverage multitask learning~\cite{simpose} or incorporate geodesic distance to learn UV maps~\cite{tan2021humangps}.
BodyMap \cite{ianina2022bodymap} incorporates the Vision Transformers to learn per-pixel image features on a continuous body surface that handles loose clothes, different hairstyles, and occlusion. 
 TemporalUV \cite{Xie_2022_CVPR} focuses on handling garments by extrapolating the initial DensePose estimates \cite{guler2018densepose} and leveraging image features obtained from an input video to obtain a UV aligned with the garment boundary.
Despite their promise, the visual artifacts persist due to a lack of understanding the underlying 3D geometry. Unlike previous approaches, we design our framework such that the resulting UV map satisfies the fundamental geometric property of isometry, which results in physically plausible re-texturing. 

\subsection{Neural UV Optimization from Videos.}

Another line of work \cite{kasten2021layered, ye2022sprites, UnwrapMosaics, Lu:2020, Jampani_2017_CVPR} resorts to a layered UV map, capturing the geometry to some degree by incorporating video decomposition \cite{Wang1994RepresentingMI, Black1991RobustDM, Darrell:1991, Brostow:1999, Kumar2005LearningLM} to optimize the UV coordinates of the foreground and background based on the observed motion.
Kasten et al. \cite{kasten2021layered} unwrap a video into a set of layered 2D atlases where for each pixel in the video, its corresponding 2D coordinate in each of the atlases is predicted.
Ye et al. \cite{ye2022sprites} proposes a global sprite image that can group the distinct motion trajectories because the collective object structure has a consistent appearance throughout time. 
While preserving the temporal coherency and maintaining some coarse UV deformations related to arm or leg movements during a sequence, these methods fall short of capturing micro deformations like wrinkles in the clothing. Furthermore, these techniques cannot represent UV mapping for an image and can only be applied to videos.
Unlike these methods, our surface normal conditioned UV map is highly sensitive to small geometric details and 3D surface deformations, which can be optimized not only for a video but also for single images.

\section{Method}
\label{sec:method}



Our goal is to obtain a continuous texture mapping
, which allows editing the appearance of dynamic garments. We leverage the geometric property of isometry to constrain the UV map in the form of partial differential equations.
We solve this partial differential equations by optimizing a neural network to generate a geometry aware UV map.


\begin{figure}[t]
\begin{center}
\vspace{-5mm}
    {\includegraphics[width=0.495\textwidth]{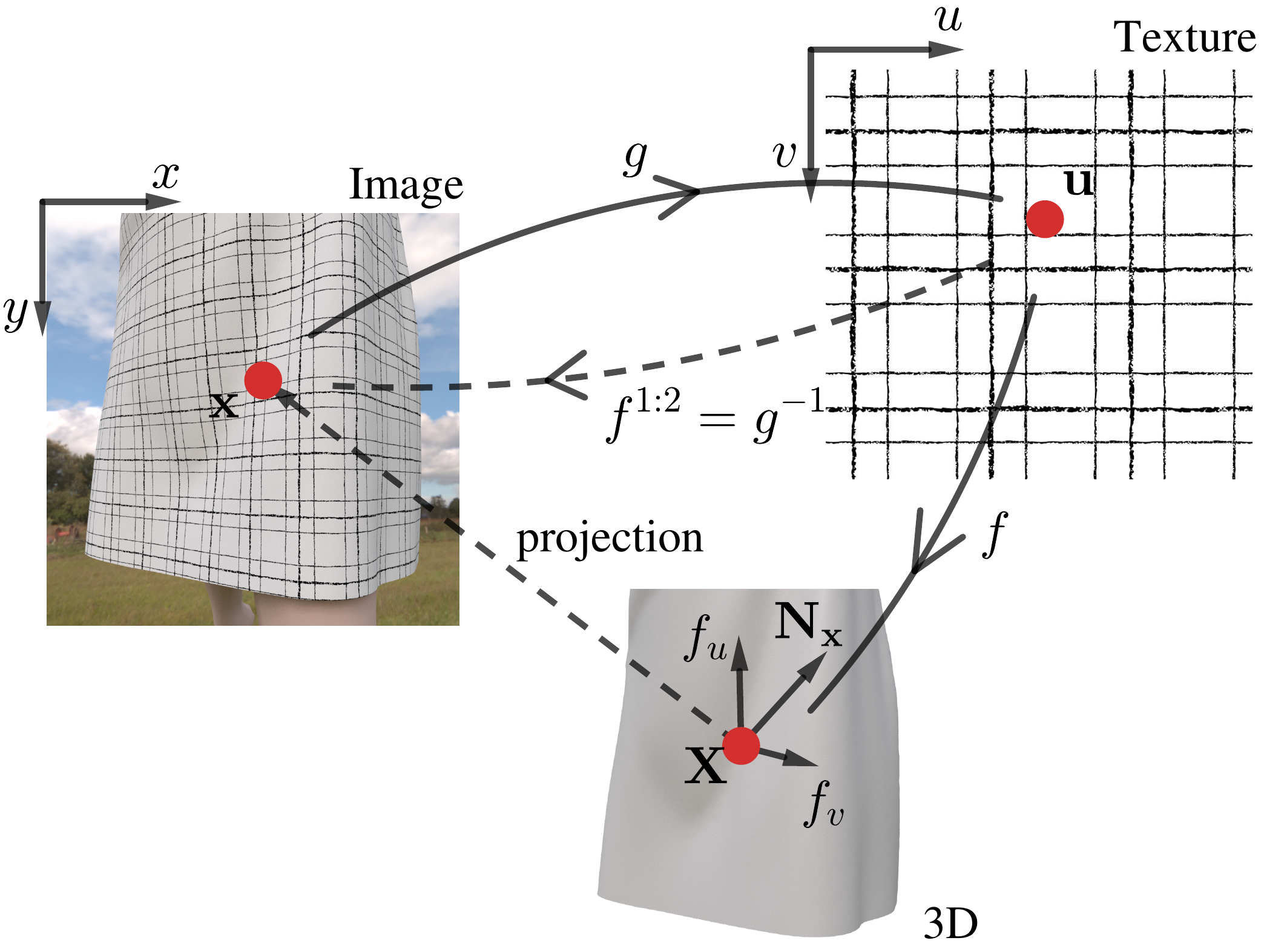}}\\\vspace{-2mm}
\end{center}
\vspace{-3mm}
   \caption{\textbf{Texture mapping geometry.} We study the mapping between the image space, texture space, and 3D space. A point $\textbf{x}$ in image is mapped to the texture space with $\textbf{u} = g(\textbf{x})$. The mapping $f(\textbf{u})=\textbf{X}$ lifts the texture plane into a 3D garment surface by an isometric warping. We use the orthographic projection model, resulting in the first two elements of $f$ is the inverse of $g$, i.e., $f^{1:2} = g^{-1}$. The spatial derivatives of $f$ form a tangent plane on the 3D surface at $\mathbf{X}$, resulting in  $\mathbf{N}_\mathbf{x} = f_u \times f_v$ where $\mathbf{N}_\mathbf{x}$ is the 3D surface normal, and $f_u$ and $f_v$ are the spatial derivatives of $f$ with respect to $u$ and $v$. 
   }
\label{Fig:illustration}
\vspace{-5mm}
\end{figure}
\subsection{Texture Mapping without 3D Reconstruction} \label{sec:map}
Consider a mapping $g(\textbf{x}) = \textbf{u}$ that maps a pixel location $\textbf{x} = (x,y) \in \mathds{R}^2$ in the image space that belongs to a garment of interest to a point $\textbf{u} = (u,v)\in \mathds{R}^2$ in the UV space of the garment as shown in Figure~\ref{Fig:illustration}. 
The goal of our work is to find such a mapping $g$ that takes into account the local surface geometry measured by the surface normal predicted at $\mathbf{x}$. 
We denote the predicted 3D surface normal of $\mathbf{x}$ in the camera space as $\mathbf{N}_\mathbf{x}\in \mathds{S}^2$. 

Let us define an \textit{isometric} map from the UV texture map to the 3D surface, $f(\mathbf{u}) = \mathbf{X}$. 
This is the fundamental property of a non-stretchable cloth texture mapping~\cite{Catmull1974ASA}.
\begin{align}
    &\left\|f_u\right\| = \left\|f_v\right\| = 1,   
    ~~f_u^\mathsf{T} f_v = 0, \label{eq:constraint} 
\end{align}
where $f_u$ and $f_v$ are the partial derivatives of $f$ with respect to $u$ and $v$, respectively. Geometrically, $f_u$ and $f_v$ are the tangential vectors on the 3D surface where their cross product forms the surface normal:
\begin{align}
    \widetilde{\mathbf{N}}_{\mathbf{x}} = {f_u} \times {f_v}= {f_u}(g(\mathbf{x})) \times {f_v}(g(\mathbf{x}))\label{Eq:cross}
\end{align}
where $\widetilde{\mathbf{N}}_{\mathbf{x}}\in \mathds{S}^2$ is the surface normal at $\mathbf{X}$ corresponding to $\mathbf{x}$. 

We can find the UV mapping $g$ by matching the surface normal $\widetilde{\mathbf{N}}_{\mathbf{x}}$ derived by Equation~(\ref{Eq:cross}) and the surface normal predicted from the image $\mathbf{N}_\mathbf{x}$:
\begin{align}
    \underset{\boldsymbol{\theta}_g,\boldsymbol{\theta}_f}{\operatorname{minimize}} &\sum_{\mathbf{x}} \left\|f_u(g(\mathbf{x}))\times f_v(g(\mathbf{x})) - \mathbf{N}_\mathbf{x}\right\|^2,\nonumber\\
    &{\rm s.t.}~~~\left\|f_u\right\| = \left\|f_v\right\| = 1, f_u^\mathsf{T} f_v = 0, \label{Eq:opt}
\end{align}
where $\boldsymbol{\theta}_g$ and $\boldsymbol{\theta}_f$ are the parameters of the function $g$ and $f$, respectively.
A key challenge of solving Equation~(\ref{Eq:opt}) lies in the dependency of $f$ 
that requires full 3D reconstruction of the surface. 
Instead, we formulate a new dual problem that can solve Equation~(\ref{Eq:opt}) effectively without finding $f$.

We use two properties to eliminate $f$ from Equation~(\ref{Eq:opt}). First, we assume orthographic projection, i.e., $(x,y) = (X,Y)$ where $\mathbf{X} = \begin{bmatrix}X & Y & Z\end{bmatrix}^\mathsf{T}$. This allows us to express the 3D derivatives using the pixel coordinates:
\begin{align}
    g = \left(f^{1:2}\right)^{-1}, 
\end{align}
where $f^{1:2}$ is the first two elements ($X,Y$) of $f$. To keep $f^{1:2}$ bijective, we assume there is no self occlusion in the camera projection of $f$. Note that $g^{-1}$ is the inverse of $g$ that maps the UV texture map to the pixel coordinate. Second, we derive the derivatives of $g$ by using the inverse function theorem~\cite{Felix_inverse_func}:
\begin{align}
f^{1:2}\circ g(\mathbf{x}) = \mathbf{x} ~~\rightarrow~~ \mathbf{J}_g = \left(\mathbf{J}_{(f^{1:2})}\right)^{-1}, \label{eq:jacobian}
\end{align}
where $\mathbf{J}_g=\begin{bmatrix}g_x & g_y\end{bmatrix} = \begin{bmatrix}\frac{\partial u}{\partial x} & \frac{\partial u}{\partial y}\\\frac{\partial v}{\partial x} & \frac{\partial v}{\partial y}
\end{bmatrix}$ is the Jacobian matrix of the function $g$. 

With Equation~(\ref{eq:constraint}) and (\ref{eq:jacobian}), Equation~(\ref{Eq:cross}) can be re-written as the following constraints by eliminating $f$ (3D reconstruction):
\begin{align}
    \hspace{-3mm}\left\|g_x\right\| = \sqrt{1+\frac{\widetilde{n}_x^2}{\widetilde{n}_z^2}},~
    \left\|g_y\right\| = \sqrt{1+\frac{\widetilde{n}_y^2}{\widetilde{n}_z^2}},~
    g_x^\mathsf{T} g_y = \frac{\widetilde{n}_x\widetilde{n}_y}{\widetilde{n}_z^2}, \label{eq:new_constraint}
\end{align}
where $\widetilde{\mathbf{N}}_{\mathbf{x}} = \begin{bmatrix}\widetilde{n}_x & \widetilde{n}_y & \widetilde{n}_z\end{bmatrix}^\mathsf{T}$.  For the derivation of Equation~(\ref{eq:new_constraint}), see Supplementary Material.

\begin{figure}[t]
\begin{center}
\vspace{-5mm}
    {\includegraphics[width=0.45\textwidth]{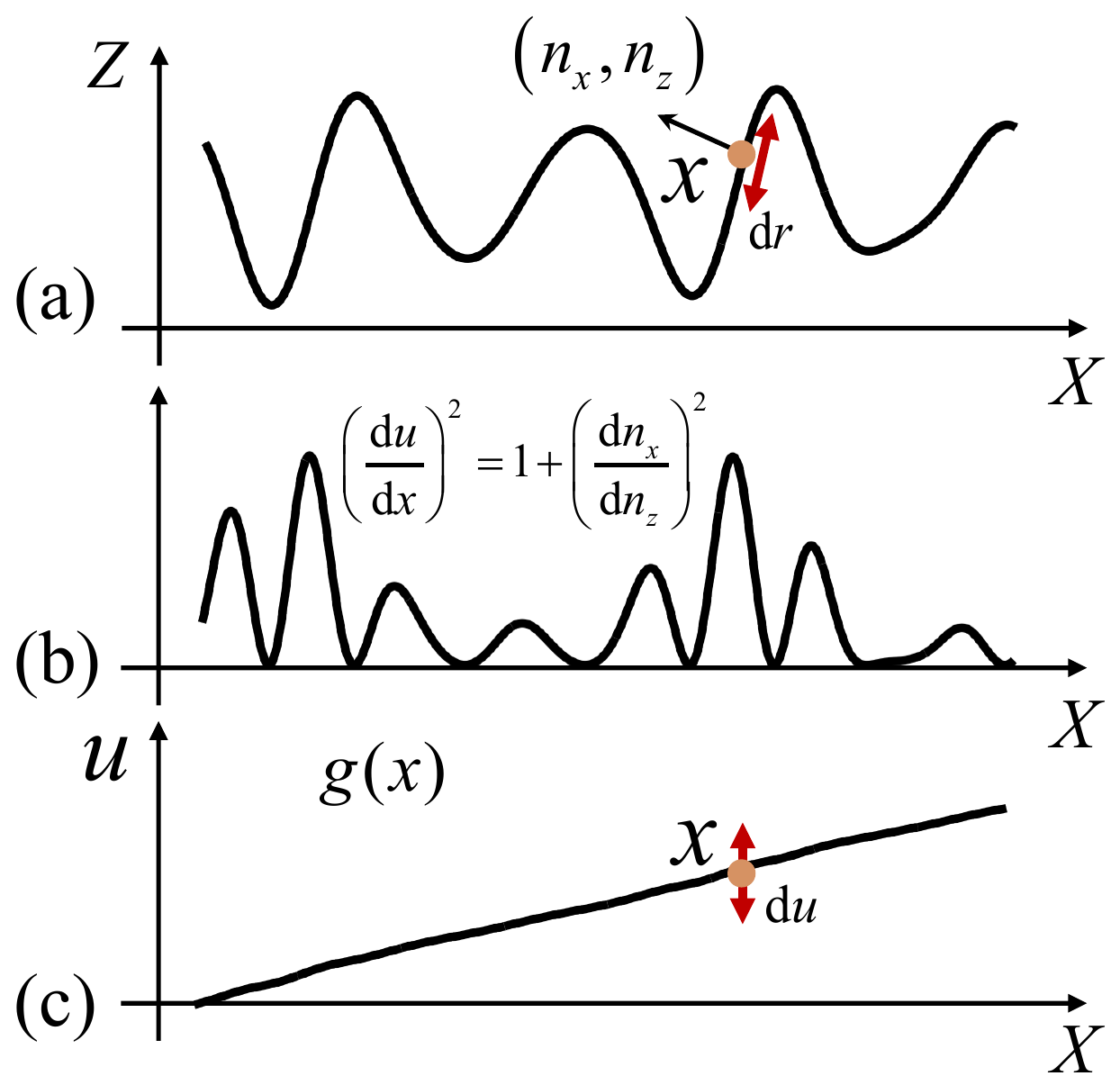}}\\\vspace{-2mm}
\end{center}
\vspace{-3mm}
   \caption{\textbf{2D simplification of UV map estimation.} We illustrate the isometric relationship between (a) the surface and (c) the UV map. The length of the curve in XZ axis needs to be preserved when mapping to $g(x)$, i.e., $|{\rm d}u| = |{\rm d}r| = \sqrt{{\rm d}x^2+{\rm d}z^2}$ (arc length preservation). This relationship can be re-written as (b) a partial differential equation using Equation~(\ref{eq:new_constraint}) in terms of its surface normal and the inverse of the spatial derivative of $g$, i.e., $\left(\frac{{\rm d}u}{{\rm d}x}\right)^2=1+\left(\frac{{\rm d}n_x}{{\rm d}n_z}\right)^2$.}
\label{Fig:geo}
\vspace{-5mm}
\end{figure}

Equation~(\ref{eq:new_constraint}) is a set of partial differential equations of $g$ that needs to match with the predicted surface normal $\mathbf{N}_\mathbf{x} = \begin{bmatrix}n_x&n_y&n_z\end{bmatrix}^\mathsf{T}$, leading to a loss function:
\begin{align}
    \hspace{-3mm}\mathcal{L}_{\rm geo} (\boldsymbol{\theta}_g) &= \sum_\mathbf{x}\left(\left(\left.\frac{\partial u}{\partial x}\right|_\mathbf{x}\right)^2+\left(\left.\frac{\partial v}{\partial x}\right|_\mathbf{x}\right)^2-1-\frac{n_x^2}{n_z^2}\right)^2 \nonumber\\
    &+ \left(\left(\left.\frac{\partial u}{\partial y}\right|_\mathbf{x}\right)^2+\left(\left.\frac{\partial v}{\partial y}\right|_\mathbf{x}\right)^2-1-\frac{n_y^2}{n_z^2}\right)^2 \nonumber\\
    &+ 
    \left(\left.\frac{\partial u}{\partial x}\right|_\mathbf{x}\left.\frac{\partial u}{\partial y}\right|_\mathbf{x}+\left.\frac{\partial v}{\partial x}\right|_\mathbf{x}\left.\frac{\partial v}{\partial y}\right|_\mathbf{x}-\frac{n_x n_y}{n_z^2}\right)^2,
    \label{eq:loss_geo}
\end{align}
where 
$\left.\frac{\partial u}{\partial x}\right|_\mathbf{x}$ is the partial derivative of $u$ with respect to $x$ evaluated at $\mathbf{x}$.

Figure~\ref{Fig:geo} illustrates a 2D simplification of UV map estimation. A curve in XZ plane forms a isometric relation with the UV map, $|{\rm d}u| = |{\rm d}r| = \sqrt{{\rm d}x^2+{\rm d}z^2}$. This relationship can be re-written as a partial differential equation in terms the surface normal $(n_x, n_z)$ and the inverse of the spatial derivative of $g(x)$ using Equation~(\ref{eq:new_constraint}), i.e., $\left(\frac{{\rm d}u}{{\rm d}x}\right)^2=1+\left(\frac{{\rm d}n_x}{{\rm d}n_z}\right)^2$. We solve these partial differential equations to estimate $g$.

\subsection{Self-supervised Learning of Texture Mapping}
The texture map $g$ is defined up to a bijective function, i.e., there exists an infinite number of $g$ that are equivalent: $g^{-1}\circ g = \left(\mathcal{T}\circ g\right)^{-1}\circ (\mathcal{T}\circ g)$, where $\mathcal{T}$ is a bijective map (e.g., Euclidean transform). We resolve this ambiguity by finding $g$ such that $g\approx g'$ where $g'$ is a pre-defined proxy map of humans:
\begin{align}
    \mathcal{L}_{\rm prox}(\boldsymbol{\theta}_g) = \sum_{\mathbf{x}} \|g'(\mathbf{x})-g(\mathbf{x})\|^2.
\end{align}
In practice, we use an extended DensePose \cite{guler2018densepose} as the pre-defined proxy map. Since DensePose makes predictions only for the human body, we apply an extrapolation method~\cite{Telea_inpaint} to inpaint the garment regions that are not covered by DensePose.

Further, we ensure physical plausibility of the visible 3D surfaces, i.e., the texture map should result in the surface normals pointing to $+Z$ direction, by adding the following loss:
\begin{align}
    \mathcal{L}_{\rm z}(\boldsymbol{\theta}_g) = \sum_\mathbf{x} \max(0, {\rm det}(\left.\mathbf{J}_g\right|_\mathbf{x})), \label{eq:determinant}
\end{align}
where ${\rm det}(\mathbf{J}_g)$ is the determinant of the Jacobian $\mathbf{J}_g$ that is equivalent to $\widetilde{n}_z$. $\mathbf{J}_g|_\mathbf{x}$ is the Jacobian matrix of $g$ evaluated at $\mathbf{x}$. See Supplementary Material for derivation.

For a video, we extend the texture map to include the image feature for each pixel, i.e., $g(\mathbf{x}, \mathbf{f}_\mathbf{x})$ where $\mathbf{f}_\mathbf{x}$ is the image feature at $\mathbf{x}$. This allows us to generalize the texture map over time. With the extension, we ensure the temporal consistency of the texture map by leveraging optical flow across frames:
\begin{align}
    \mathcal{L}_{\rm tmp}(\boldsymbol{\theta}_g) = \sum_{i,j}\sum_{\mathbf{x}_i} \left\|g(\mathbf{x}_i, \mathbf{f}_{\mathbf{x}_i}) - g\left(\mathbf{x}_j, \mathbf{f}_{\mathbf{x}_j}\right)\right\|^2,
\end{align}
where $\mathbf{x}_i$ is a point in the $i^{\rm th}$ frame. This point is mapped to $\mathbf{x}_j$ in the $j^{\rm th}$ frame, i.e., $\mathbf{x}_j = W_{i\rightarrow j}(\mathbf{x}_i)$ where $W_{i\rightarrow j}$ is the optical flow from the $i^{\rm th}$ to $j^{\rm th}$ frames.

Overall, we optimize the following loss to learn the texture map:
\begin{align}
    \mathcal{L}(\boldsymbol{\theta}_g) = \mathcal{L}_{\rm geo} + \lambda_{\rm prox}\mathcal{L}_{\rm prox} + \lambda_{\rm z}\mathcal{L}_{\rm z} + \lambda_{\rm tmp} \mathcal{L}_{\rm tmp},
\end{align}
where $\lambda_{\rm prox}$, $\lambda_{\rm z}$, and $\lambda_{\rm tmp}$ are the weights that determine the relative importance of losses. Note that when a single image is used, $\lambda_{\rm tmp} = 0$.

\subsection{Implementation Details}
\label{chapter:framework}

\begin{figure}[t]
\begin{center}
    {\includegraphics[width=0.495\textwidth]{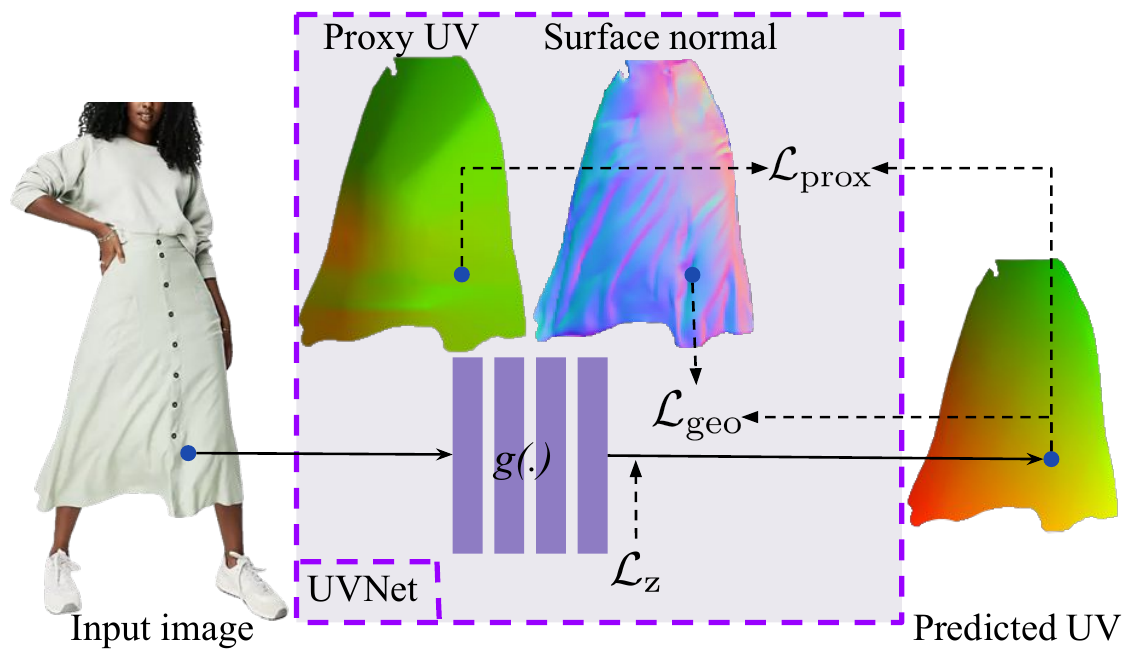}}\\\vspace{-2mm}
\end{center}
\vspace{-3mm}
   \caption{\textbf{Single image framework.} For each pixel location on a garment $\mathbf{x}$, we predict the UV coordinate $\mathbf{u}$ using a multi-layer perceptron. We enforce isometry to the UV map by matching with the predicted 3D surface normals. $\mathcal{L}_{\rm z}$ ensures that the predicted surface normals point toward the camera. To disambiguate the frame of reference of the UV map, we use pre-defined proxy map (e.g., DensePose~\cite{guler2018densepose} extrapolation \cite{Telea_inpaint}) by applying $\mathcal{L}_{\rm prox}$.}
\label{Fig:network}
\vspace{-5mm}
\end{figure}

\begin{figure*}[h]
\begin{center}
\vspace{-10mm}
    {\includegraphics[width=\textwidth]{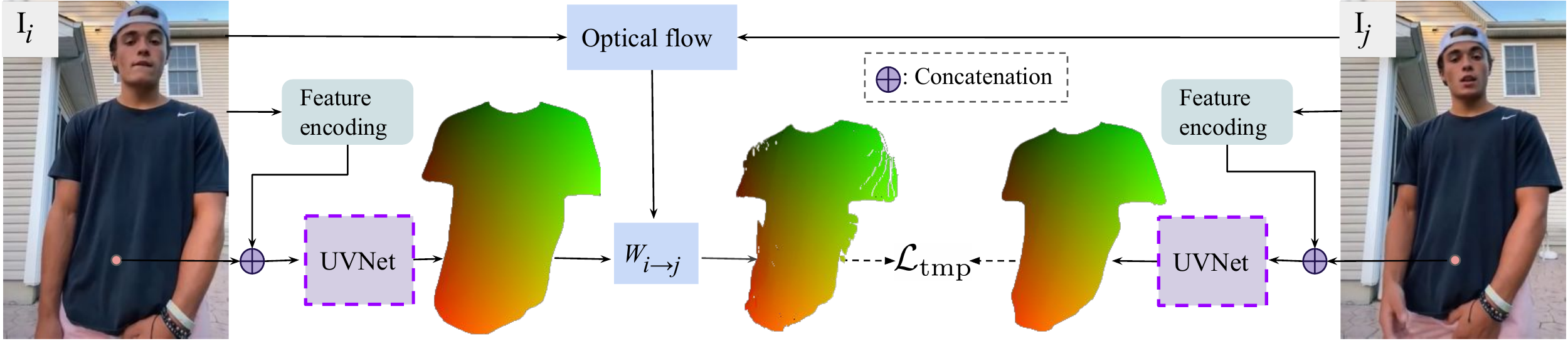}}
\end{center}
\vspace{-3mm}
   \caption{\textbf{Video framework.} We apply temporal coherence in the predicted UV maps using optical flow \cite{LucasK81, LucasK85, raft_2020}. With the optical flow $W_{i\rightarrow j}$, we match the UV prediction of $i^{\rm th}$ frame with $j^{\rm th}$ via $\mathcal{L}_{\rm tmp}$.
   }
\label{Fig:networkvid}
\end{figure*}

We model $g$ using a 12-layer multi-layer perception, with ReLU \cite{relu} as an activation function after each layer that takes as input a pixel coordinate with positional encoding,  $\gamma(\mathbf{x})$ where $\gamma:\mathds{R}^2\rightarrow\mathds{R}^{128}$ is Fourier based positional encoding~\cite{tancik2020fourfeat}. 
For videos, we use ResNet~\cite{He_2016_CVPR_resnet} to extract per-frame 256 dimensional image feature $\mathbf{f}_\mathbf{x}$. Our network design is illustrated in Figure \ref{Fig:network} (for image-based UV map prediction) and \ref{Fig:networkvid} (for video-based UV map prediction). To make our prediction scale-invariant, we crop the garments region with $256\times256$ resolution. We use an off-the-shelf garment segmentation software, Graphonomy \cite{Gong2019Graphonomy} to separate the garment area. We use Adam optimizer \cite{kingma:2015} with batch size of $2048$ and learning rate of $10^{-4}$. 
We set  $\lambda_{\rm prox} = 0.2$, $\lambda_{\rm z} = 0.01$, and $\lambda_{\rm tmp} = 0.3$.
We used an NVIDIA
V100 GPU and Intel(R) Xeon(R) CPU E5-2698 v4 @ 2.20GHz, and implemented our approach
with Pytorch \cite{pytorch}. 
Our method takes 20 minutes for a video of 82 frames while Kasten et al. \cite{kasten2021layered}, Ye et al. \cite{ye2022sprites}, and TemporalUV \cite{Xie_2022_CVPR} take 10 hours, 30 minutes, and 23 hours, respectively.

\section{Evaluation}
\label{sec:evaluation}

We evaluate our method both quantitatively and qualitatively on real images as well as synthetic data with ground truth UV map. We also compare with the state-of-the-art methods on human UV map estimation methods and video UV map optimization approaches.

\begin{table*}[t]
  \small
  \centering
  \begin{tabular}{l@{\hskip 0.1in}l@{\hskip 0.1in}l@{\hskip 0.1in}l@{\hskip 0.1in}l@{\hskip 0.1in}l@{\hskip 0.1in}l@{\hskip 0.1in}}
     \toprule                          
       &  ~~~~~~~~~~~~GT dress&\hspace{-2mm}sequences~\cite{santesteban2021garmentcollisions}   & ~~~~~~~~~~~~GT T-shirt&\hspace{-2mm}sequences~\cite{santesteban2021garmentcollisions} & ~~~~~~~~Real Fashion&\hspace{-2mm}sequences~\cite{Zablotskaia2019DwNetDW} \\
      \cmidrule(r){1-1} \cmidrule(r){2-3} \cmidrule(r){4-5} \cmidrule(r){6-7}
    Method                    & UV. error (cm)           &  photo. error     & UV. error (cm)           &  photo. error  & geo. error ($\mathcal{L}_{\rm geo}$)            &  tmp. error  \\
    \cmidrule(r){1-1}\cmidrule(r){2-2} \cmidrule(r){3-3} \cmidrule(r){4-4} \cmidrule(r){5-5} \cmidrule(r){6-6} \cmidrule(r){7-7}
DensePose \cite{guler2018densepose}    &  17.27$\pm$3.76 & 52.43$\pm$10.20                               & 7.48$\pm$0.52   &  34.19$\pm$7.01  & 1.26$\pm$0.47 & 8.12$\pm$2.70\\
Extrapolated DensePose \cite{guler2018densepose,Telea_inpaint}   &  8.61$\pm$0.76  &  18.29$\pm$3.24     & 5.34$\pm$0.63   &  18.86$\pm$4.55  & 0.52$\pm$0.06 & 4.62$\pm$0.73   \\   
HumanGPS \cite{tan2021humangps}         & 11.97$\pm$2.06   &  97.41$\pm$30.87                            & 7.53$\pm$0.43   & 108.10$\pm$34.89 & 1.27$\pm$0.69 & 51.40$\pm$41.66  \\
Kasten et al. \cite{kasten2021layered}         & 7.06$\pm$0.60     & 13.07$\pm$2.79                      & 6.64$\pm$0.63    &  14.54$\pm$4.26  & 0.56$\pm$0.10  & 2.30$\pm$0.70\\
Ye et al. \cite{ye2022sprites}         & 5.56$\pm$0.29             & 33.57$\pm$-10.27                    & 5.75$\pm$0.20    &  19.22$\pm$5.14  & 0.71$\pm$0.03& 1.65$\pm$0.15 \\
\cmidrule(r){1-1}\cmidrule(r){2-2} \cmidrule(r){3-3} \cmidrule(r){4-4} \cmidrule(r){5-5}  \cmidrule(r){6-6} \cmidrule(r){7-7}
{Ours}         & \textbf{3.16$\pm$0.28}   & \textbf{7.54$\pm$2.04}                                       &\textbf{3.58$\pm$0.27} &\textbf{11.28$\pm$2.01} & \textbf{0.07$\pm$0.03} & \textbf{1.50$\pm$0.23}\\
    \bottomrule
  \end{tabular}
  \vspace{1mm}
  \vspace{-2mm}
  \caption{Quantitative Results. UV. error (cm), photo. error (RGB difference), geo. error ($\mathcal{L}_{\rm geo}$), and tmp. error ($\mathcal{L}_{\rm tmp}$) (image space pixel distance) respectively (mean$\pm$std).
  }\label{table:Quant1}
\vspace{-7mm}
\end{table*}

\begin{table}[t]
  \small
  \centering
  \vspace{-0mm}
  \begin{tabular}{l@{\hskip 0.1in}l@{\hskip 0.1in}l@{\hskip 0.1in}}
     \toprule                           
    Method                    & UV. error (cm)           &  photo. error (RGB)   \\
    \cmidrule(r){1-1}\cmidrule(r){2-2} \cmidrule(r){3-3} 
{Ours}                                                  & \textbf{3.16$\pm$0.28}    & \textbf{7.54$\pm$2.04}  \\
{Distance constraint}               & 3.44$\pm$0.25             & 8.29$\pm$1.99   \\
{Angle constraint }               & 5.96$\pm$0.33             & 7.68$\pm$1.37   \\
{W/o $\mathcal{L}_{\rm tmp}$}                        & 3.22$\pm$0.24             & 14.67$\pm$4.82   \\
{W/o $\mathcal{L}_{\rm prox}$}                       & 3.24$\pm$0.35         & 8.46$\pm$2.19\\ 
    \bottomrule
  \end{tabular}
  \vspace{1mm}
  \vspace{-2mm}
  \caption{Ablation study on dress sequences~\cite{santesteban2021garmentcollisions}. UV. error (cm), photo. error (RGB difference) (mean$\pm$std). }
  \label{table:ablation}
\end{table}

\noindent\textbf{Evaluation Datasets}
We evaluate our method using the following datasets: (1) five synthetic video sequences of simulated dress and T-shirt garments 
from Santesteban et al.~\cite{santesteban2021garmentcollisions}
with random texture patterns over 700 frames;
(2) ten real videos from Fashion Video dataset \cite{Zablotskaia2019DwNetDW};
(3) TikTok dataset \cite{Jafarian_2021_CVPR_TikTok} and various YouTube videos as well as in-the-wild internet images.

\begin{figure}[t]
  \begin{center}
  \vspace{-0mm}
    \includegraphics[width=0.45\textwidth]{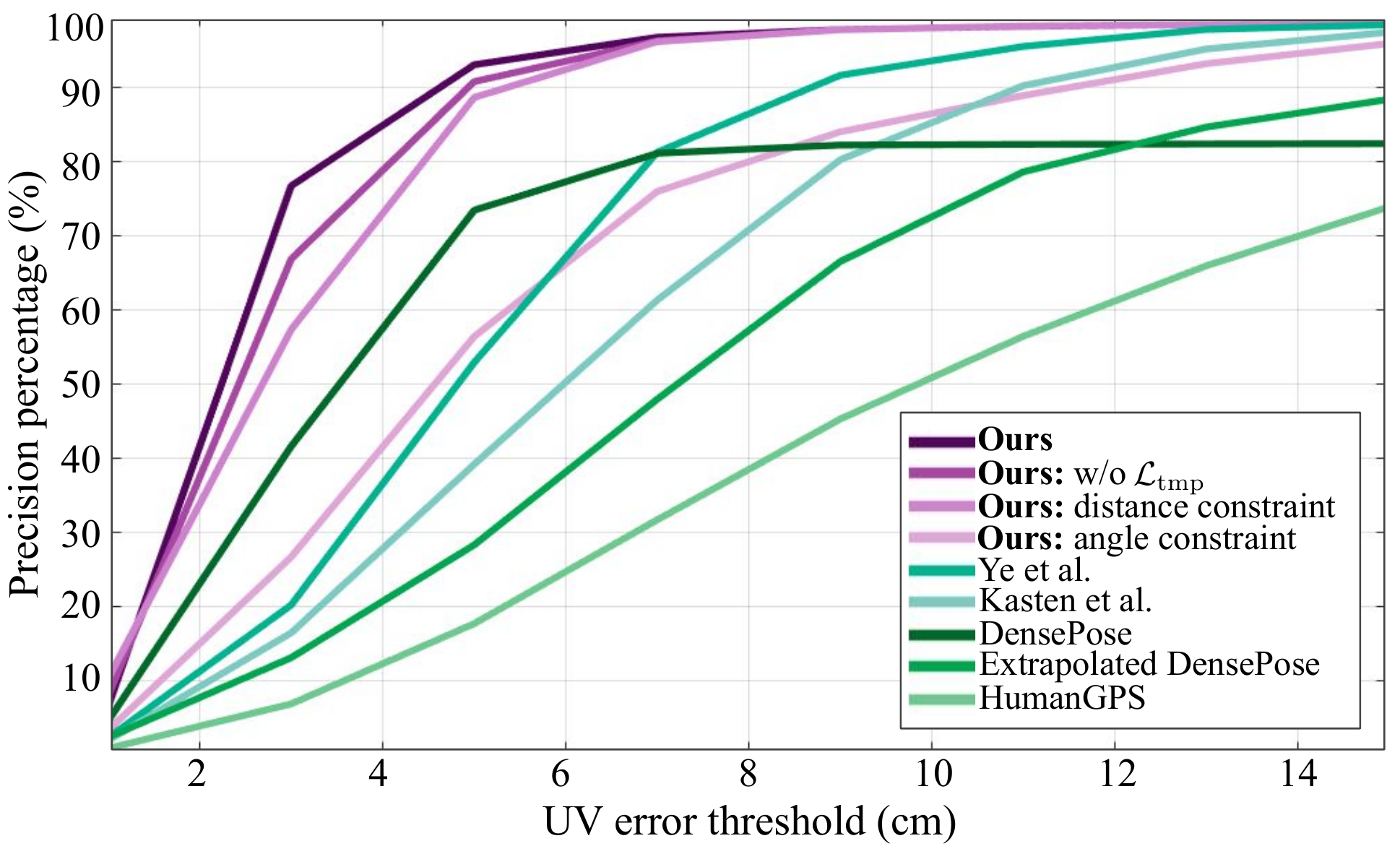}
  \end{center}
     \vspace{-8mm}
  \caption{\textbf{Average precision.} We compute the average of pixels with a UV error higher than a threshold (1-15cm) on our method, the baseline methods, and our ablation experiments on ground truth dress sequences~\cite{santesteban2021garmentcollisions}.}
  \label{Fig:AP}
  \vspace{-8mm}
\end{figure}

\noindent\textbf{Evaluation Metric} 
We use five metrics to evaluate our method.
(1) UV error: for synthetic data,
we report the absolute UV error in the texture space~\cite{santesteban2021garmentcollisions}. We use a Procrustes analysis \cite{Procrustes_2005} to align the resulting texture map to account for the diambiguity of the reference frame. We report the mean squared error in metric scale by assuming the height of the person in the input is 165cm, resulting in 0.41cm/UV for dress and 0.27cm/UV for T-shirt (Table \ref{table:Quant1}).\\ 
(2) Average Precision percentage: we report the Average Precision (AP) percentage metric computed on all the pixels considering a per-pixel prediction as correct if the UV error is lower than a threshold.
We visualize the AP metric for a range of thresholds from 1 to 15 cm and obtain the graph shown in Figure \ref{Fig:AP}.\\
(3) Photometric error: we warp the first frame of an input video to the rest of the frames using the UV map estimates. We report the error between the ground truth RGB images and the warped RGB images as reported in Table \ref{table:Quant1}.\\
(4) Geometric error: we report the $\mathcal{L}_{\rm geo}$ to show how the predicted UV map follows the geometric information captured in the surface normal estimates as reported in Table \ref{table:Quant1}. \\
(5) Temporal error: we evaluate the capability of the different approaches in preserving the temporal coherency by reporting the  $\mathcal{L}_{\rm tmp}$ error (Table \ref{table:Quant1}).\\
When the ground truth UV is not available, we use the geometric and temporal errors to evaluate our method.
%

\begin{figure}[!h]
\vspace{-4mm}
\begin{center}
    {\includegraphics[width=0.486\textwidth]{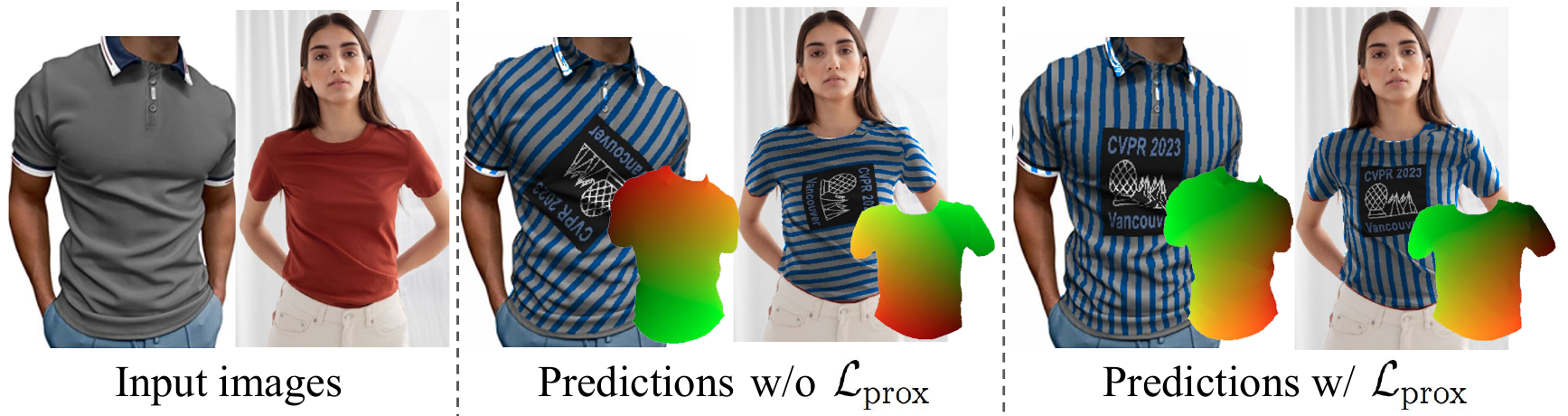}}\\\vspace{-2mm}
\end{center}
\vspace{-5mm}
   \caption{The impact of $\mathcal{L}_{\rm prox}$ in UV prediction and retexturing.}
\label{Fig:Lprox}
\vspace{-5mm}
\end{figure}

\noindent\textbf{Baseline Methods} 
We compare our method with previous works that fall into two categories: (1) human UV map prediction; (2) UV optimization.

\noindent 1) \textit{Human UV map prediction:} 
we compare our method with state-of-the-art that focus on predicting UV maps for the naked human body \cite{guler2018densepose} and dressed humans \cite{tan2021humangps, Xie_2022_CVPR}. We also report the performance of the UV map obtained by extrapolating DensePose predictions as discussed in Section \ref{chapter:framework}. 
Our method achieves the best performance as shown in Table \ref{table:Quant1} and Figure \ref{Fig:AP}. We notice that DensePose \cite{guler2018densepose} performs competently in precision percentage when the threshold error is less than 7 cm. This observation is based on the fact that, for each pixel, DensePose predicts a part label (among 24 parts) and a UV map with respect to that body part. When aligning these predictions with the ground truth, we warp each occupied body part individually, resulting in a more accurate alignment compared to the other methods (including ours) that are represented by only one patch. However, the performance of DensePose \cite{guler2018densepose} is not improved above 7cm because of limited ability to predict beyond body surface. 

\noindent 2) \textit{UV optimization:} 
we compare our method with state-of-the-art in predicting the UV map of a dynamic object observed in a video \cite{kasten2021layered,ye2022sprites}. However, such methods are not tailored for garments that undergo highly non-linear transformations as the body moves. Hence, as reported in Table \ref{table:Quant1} and Figure \ref{Fig:AP}, our method surpasses these baselines in the dense UV error, the average precision percentage, the geometric error, and the temporal error.\\

\begin{figure*}[th]
\begin{center}
\vspace{-12mm}
    {\includegraphics[width=0.95\textwidth]{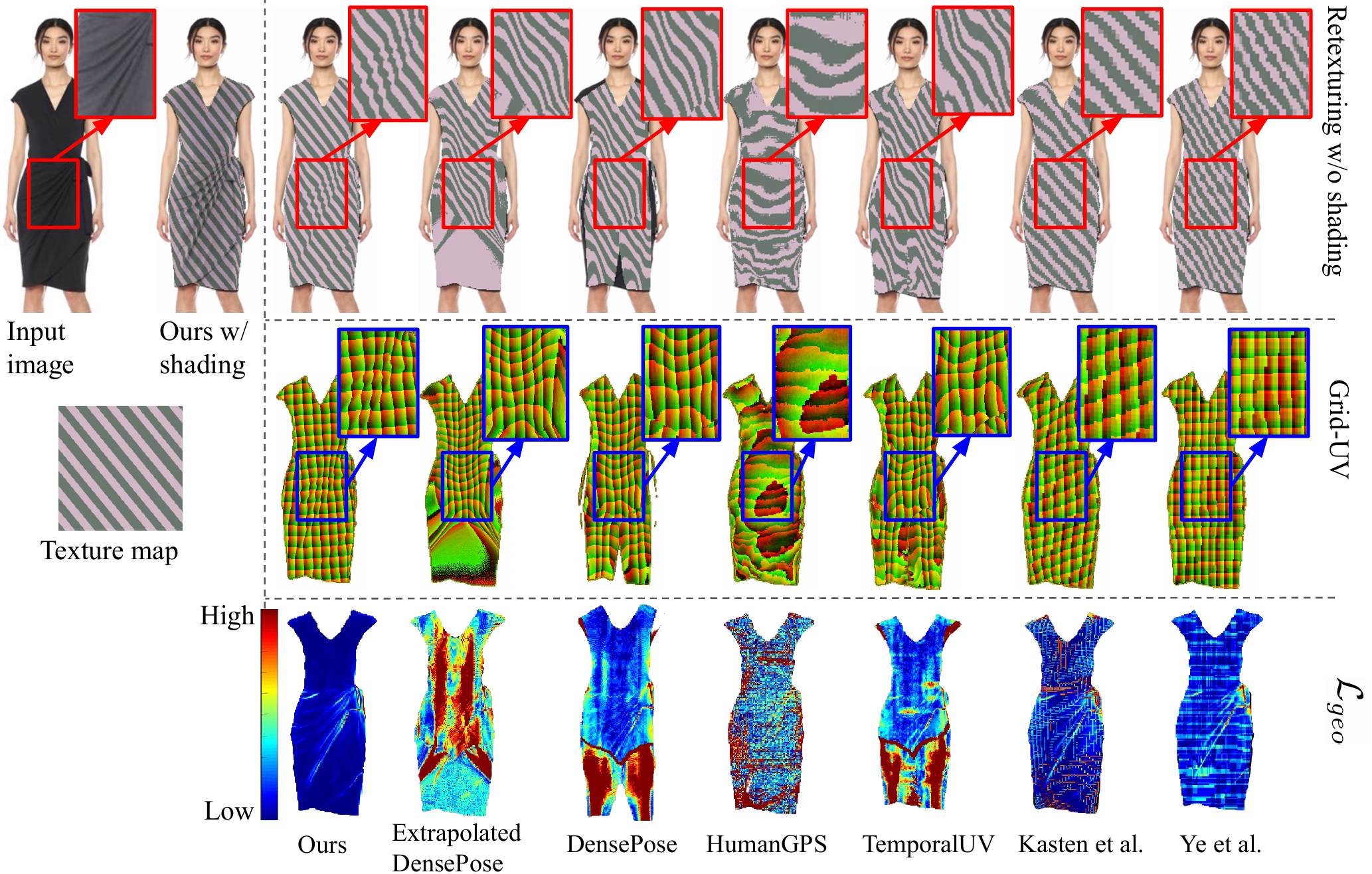}}\\\vspace{-2mm}
\end{center}
\vspace{-5mm}
   \caption{\textbf{Qualitative comparison.} We compare ours with  our initial UV (extrapolated DensePose), DensePose \cite{guler2018densepose}, HumanGPS \cite{tan2021humangps}, Kastenn et al. \cite{kasten2021layered}, and Ye et al. \cite{ye2022sprites} in image re-texturing and UV map grid visualization.}
\label{Fig:qual_compare}
\vspace{-7mm}
\end{figure*}

\begin{figure}[h]
  \begin{center}
    \includegraphics[width=0.48\textwidth]{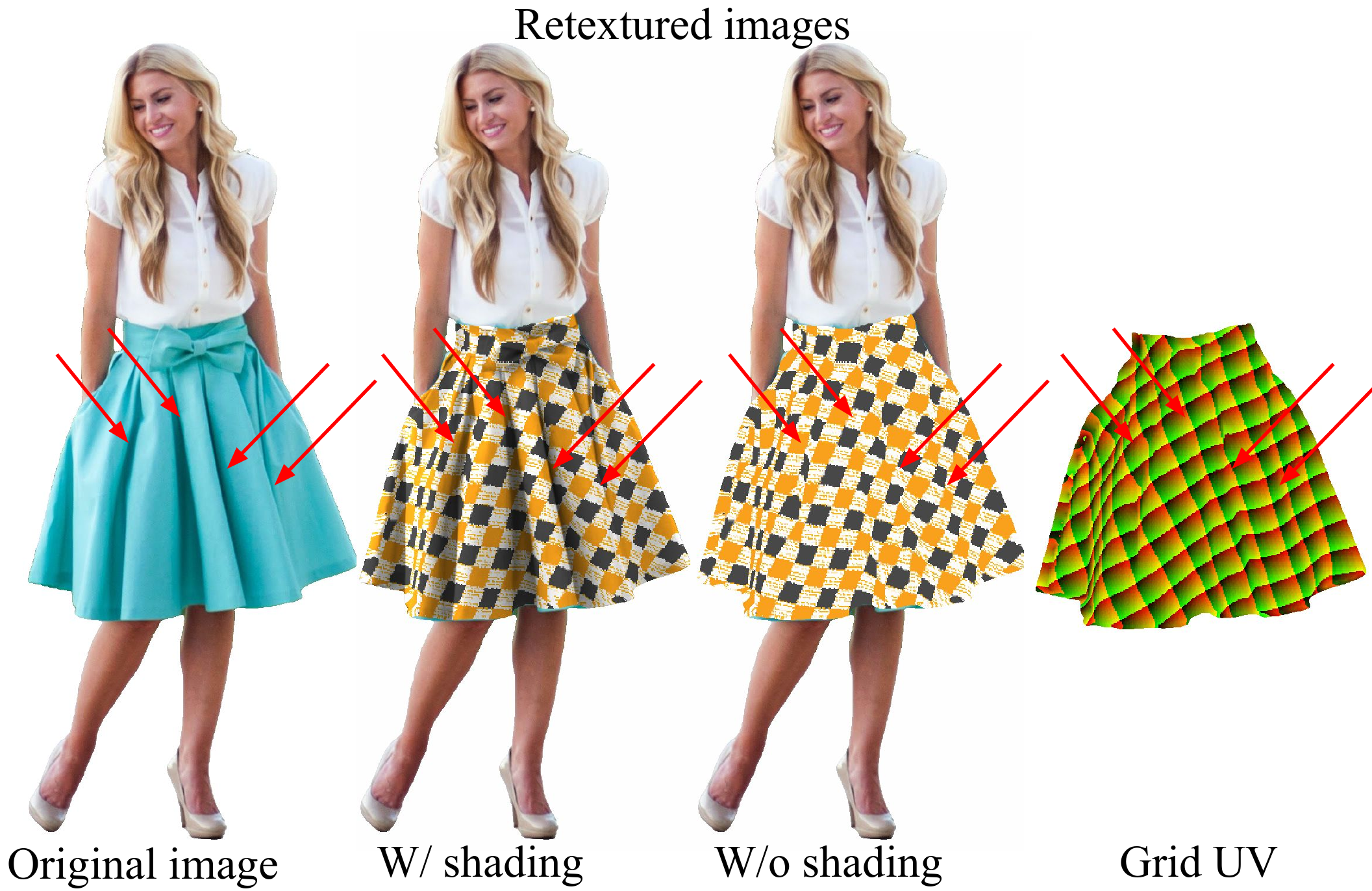}
  \end{center}
     \vspace{-5mm}
  \caption{\textbf{Limitation.} Due to lack of 3D reconstructed surface, our method cannot handle folds that introduce texture discontinuity. Our texture map is continuous around the folds, which is physically incorrect.}
  \label{Fig:limit}
  \vspace{-7mm}
\end{figure}
\vspace{-5mm}
\noindent\textbf{Ablation Study} We conduct an ablation study to analyze the impact of the distance (first two terms in Equation~(\ref{eq:loss_geo})) and angle (third term in Equation~(\ref{eq:loss_geo})) constraints (second and third rows of Table \ref{table:ablation} and Figure \ref{Fig:AP}). Our final method has the best performance in UV error and photometric error. We also compare the performance of our method without the temporal consistency ($\mathcal{L}_{\rm tmp}$) (fourth row of Table \ref{table:ablation}). As expected, this term performs quite similarly to ours in UV error but very poorly in photometric error since the consistency between the frames is not enforced.
The role of $\mathcal{L}_{\rm prox}$ is the disambiguation of UV maps, i.e., there exist an infinite number of equivalent UV maps that minimize our $\mathcal{L}_{\rm geo}$ (PDE). 
While the result without $\mathcal{L}_{\rm prox}$ are, therefore, quantitatively competitive as summarized in Table \ref{table:ablation}, such ambiguity can be resolved by finding UV that is closest to the proxy UV as shown in Figure~\ref{Fig:Lprox}, i.e., the retexture without $\mathcal{L}_{\rm prox}$ can result in arbitrary orientation across subjects.

\begin{figure*}[th]
  \centering
  \begin{subfigure}{1.0\linewidth}
  \begin{center}
    {\includegraphics[width=\textwidth]{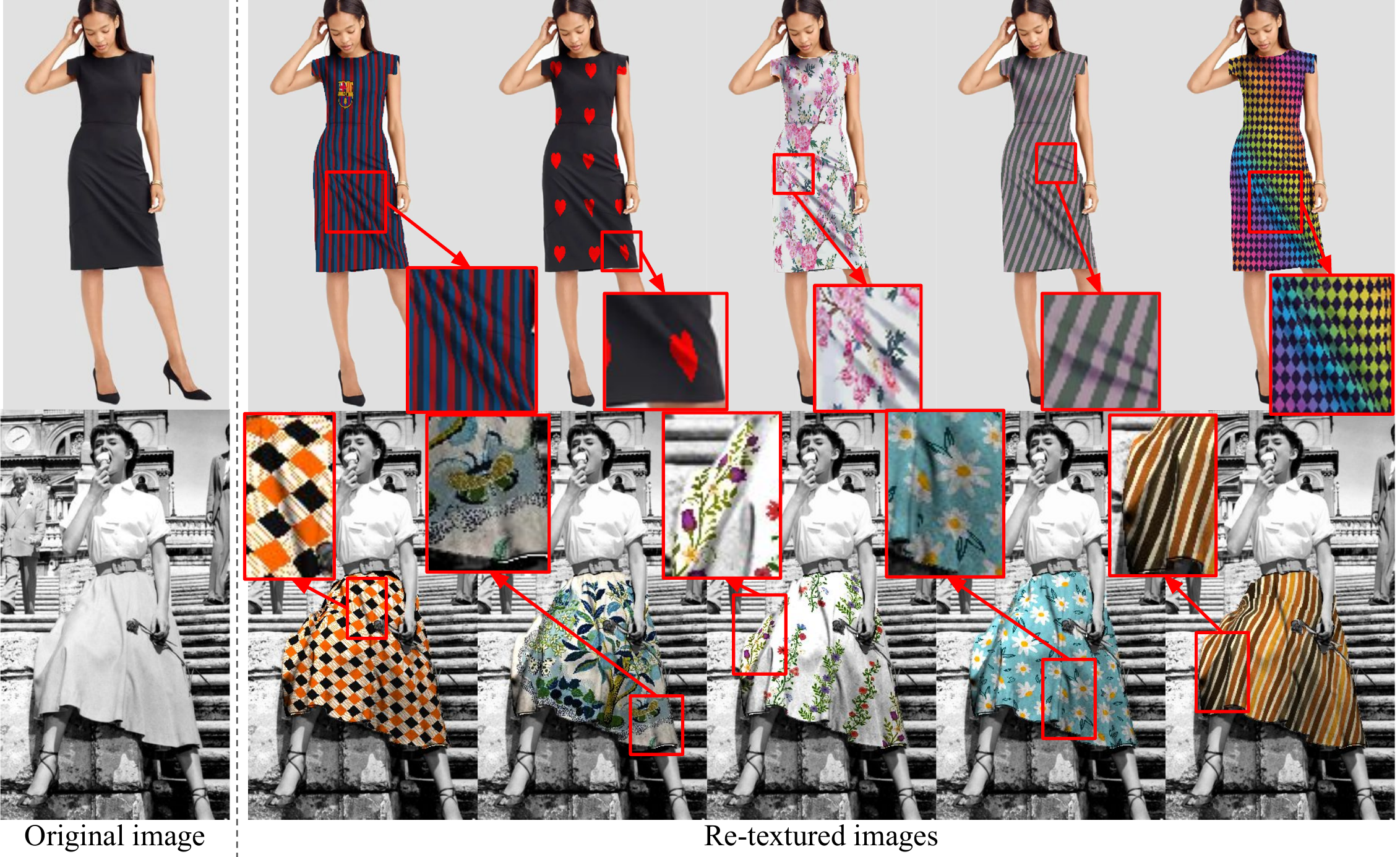}}
    \caption{Qualitative results of our method on an image.}
    \label{Fig:image_qual}
  \end{center}
  \end{subfigure}
\begin{subfigure}{1.0\linewidth}
   \begin{center}
    {\includegraphics[width=\textwidth]{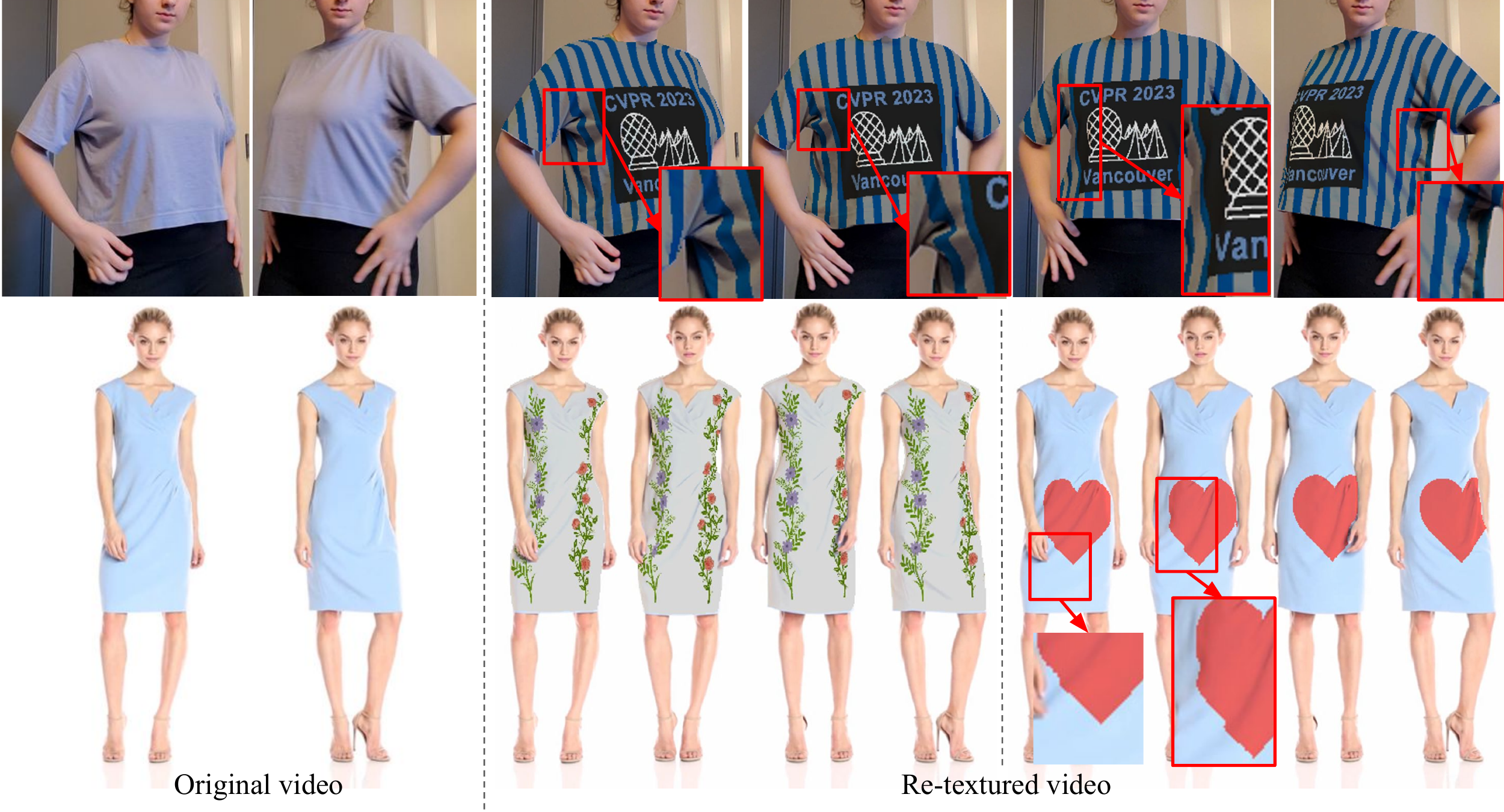}}
    \caption{Qualitative results of our method on videos.}
    \label{Fig:video_qual}
     \end{center}
  \end{subfigure}
  \caption{\textbf{Results on images and videos.} Our method can produce compelling texture editing given images and videos that preserve geometric details.
  }
  \label{fig:qual_all}
\end{figure*}

\noindent\textbf{Qualitative Results}
To show our results qualitatively, we visualize both re-texturing examples and grid UV illustration to depict the performance of each method in preserving high-frequency details. For retexturing, we first obtain an albedo and shading layer from the input image using an intrinsic image decomposition method~\cite{Weiss2001DerivingII}. After applying a new texture pattern to the albedo layer, we composite it back with the original shading layer. We apply Gamma-correction~\cite{McREYNOLDS_gamma} on the input image, after generating the albedo layer and shading layer, we inverse the Gamma-correction back when synthesizing the re-textured image. 
We compare our method qualitatively with the baselines as shown in Figure \ref{Fig:qual_compare} that illustrates the results on Fashion video sequence \cite{Zablotskaia2019DwNetDW}. Figure \ref{fig:qual_all} shows the performance of our method on videos and images. Our method not only captures the fine-grained surface details but also is temporally coherent across time.

\begin{wraptable}{r}{3.8cm}
  \footnotesize
  \centering
 \hspace{-8mm} 
 \begin{tabular}[width=0.34\textwidth]{l@{\hskip 0.1in}l@{\hskip 0.1in}}
     \toprule                           
    Method                    & User score         \\
    \cmidrule(r){1-1}\cmidrule(r){2-2} 
DensePose  &    4.09$\pm$1.94 \\
Proxy DensePose  &    3.23$\pm$1.86 \\
HumanGPS     &   2.28$\pm$2.12 \\
{Kasten et al.}       &   5.52$\pm$1.88 \\
{Ye et al.}            &  7.66$\pm$1.27     \\
{Ours}                  &  \textbf{9.57$\pm$0.81}    \\
    \bottomrule
  \end{tabular}
  \vspace{1mm}
  \vspace{-3mm}\caption{ User study. }
  \label{table:userstudy}
\vspace{-4mm}
\end{wraptable}

\vspace{-1mm}
\noindent\textbf{User Study} 
We conducted a user study: asking participants ($n=21$) to rate realism (1: unrealistic to 10: most realistic) for our method compared to the baselines as summarized in Table \ref{table:userstudy}. Our method receives the highest score from the users.

\section{Discussion}
This paper presents a novel approach to predict a high quality UV map by preserving geometric details from images and videos. We leverage the geometric property of isometry encoded in 3D surface normals to optimize the UV map in the form of partial differential equations. 
We generalize our method to videos by integrating optical flow, resulting in a temporally coherent video editing. 
Our method produces strong qualitative and quantitative predictions on real-world imagery compared to state-of-the-art UV map estimation.

\noindent\textbf{Limitation} As discussed in Section~\ref{sec:map}, our method makes an assumption about projection, i.e., there is one-to-one correspondence between 3D surface geometry and image. However, this assumption does not hold when there is a fold where a region of 3D surface is not visible to the image. This makes a contrast with 3D reconstruction based method where the invisible part of 3D surface can be still mapped to the image via depth reasoning. Figure~\ref{Fig:limit} illustrates this limitation where there are folds in the skirt, resulting in negative surface normal $n_z$. Due to the folds, the texture must be discontinuous while our method produces continuous texture rendering due to the lack of 3D reconstructed geometry, which is physically incorrect.     
\begin{wrapfigure}{r}{0.24\textwidth}
\begin{center}
\vspace{-10mm}
\hspace{-8mm}
    {\includegraphics[width=0.28\textwidth]{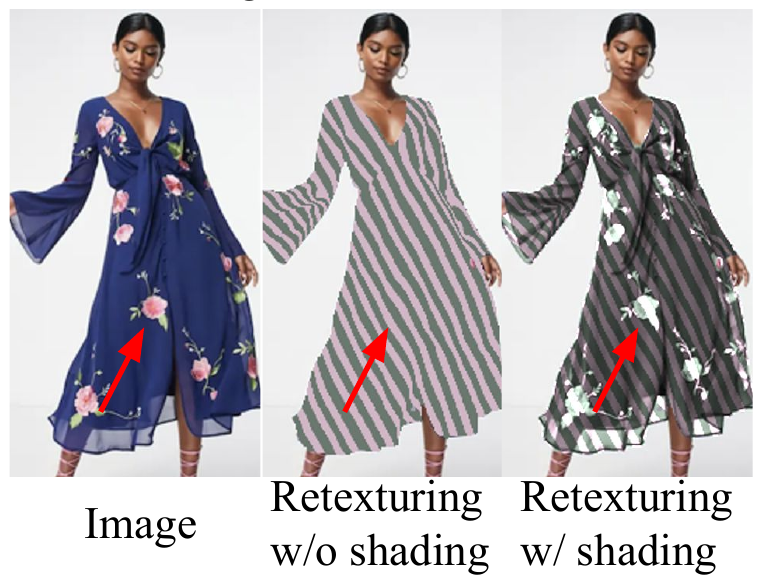}}\\\vspace{-2mm}
\end{center}
\vspace{-3mm}
   \caption{\textbf{Limitation.} Limitations on textured garment.}
\label{Fig:limit2}
\vspace{-5mm}
\end{wrapfigure}
Our pipeline is composed of two components: (1) UV prediction (our contribution) and (2) texture map with shading
(not our contribution). For the garments with highly contrasted texture, the shading operation is often biased to color contrast, resulting in erroneous appearance. 
Figure~\ref{Fig:limit2} illustrates that despite reasonable UV prediction from our method, the resulting appearance is unrealistic near the textured region. Improving the shading operation is beyond the scope of this work, and we leave it as future work.

\noindent\textbf{Acknowledgement}
This work was supported by a NSF NRI 2022894 and NSF  CAREER 1846031.

\newpage

{\small
\bibliographystyle{ieee_fullname}
\bibliography{egbib}
}

\newpage
\newpage

\section{Derivation of Equation (\ref{eq:new_constraint})}
Given Equation (2), $\widetilde{\mathbf{N}}_{\mathbf{x}} = {f_u} \times {f_v}$, the surface normal can be expressed as:
\begin{align}
    \widetilde{n}_x &= \frac{\partial y}{\partial u}\frac{\partial z}{\partial v}-\frac{\partial y}{\partial v}\frac{\partial z}{\partial u} \nonumber\\
    \widetilde{n}_y &= \frac{\partial z}{\partial u}\frac{\partial x}{\partial v}-\frac{\partial z}{\partial v}\frac{\partial x}{\partial u}  \nonumber\\
    \widetilde{n}_z &= \frac{\partial x}{\partial u}\frac{\partial y}{\partial v}-\frac{\partial x}{\partial v}\frac{\partial y}{\partial u}, \label{Eq:normal}
\end{align}
where $f_u = \begin{bmatrix}\frac{\partial x}{\partial u} & \frac{\partial y}{\partial u} & \frac{\partial z}{\partial u}\end{bmatrix}^\mathsf{T}$, $f_v = \begin{bmatrix}\frac{\partial x}{\partial v} & \frac{\partial y}{\partial v} & \frac{\partial z}{\partial v}\end{bmatrix}^\mathsf{T}$, and $\widetilde{\mathbf{N}}_{\mathbf{x}} = \begin{bmatrix}\widetilde{n}_x &\widetilde{n}_y & \widetilde{n}_z\end{bmatrix}^\mathsf{T}$.

Given the expression of the surface normal, it is possible to express $g_x$ and $g_y$ in terms of $f_u$ and $f_v$ using the inverse function theorem (Equation (5)), $\mathbf{J}_g = \left(\mathbf{J}_{f^{1:2}}\right)^{-1}$:
\begin{align}
     \begin{bmatrix}\frac{\partial u}{\partial x} & \frac{\partial u}{\partial y}\\ \\\frac{\partial v}{\partial x} & \frac{\partial v}{\partial y}
\end{bmatrix} 
&= \begin{bmatrix}\frac{\partial x}{\partial u} & \frac{\partial x}{\partial v}\\ \\\frac{\partial y}{\partial u} & \frac{\partial y}{\partial v}
\end{bmatrix} ^{-1}\\
&= \frac{1}{\frac{\partial x}{\partial u}\frac{\partial y}{\partial v} - \frac{\partial x}{\partial v}\frac{\partial y}{\partial u}} \begin{bmatrix}\frac{\partial y}{\partial v} & -\frac{\partial x}{\partial v} \\\\-\frac{\partial y}{\partial u} & \frac{\partial x}{\partial u} 
\end{bmatrix} \\ 
&=\frac{1}{\widetilde{n}_z} \begin{bmatrix}\frac{\partial y}{\partial v} & -\frac{\partial x}{\partial v} \\\\-\frac{\partial y}{\partial u} & \frac{\partial x}{\partial u}. 
\end{bmatrix} 
\end{align}
Therefore, the partial derivatives of $g$ with respect to $x$ and $y$ can be written as:
\begin{align}
    g_x = \frac{1}{\widetilde{n}_z}\begin{bmatrix}\frac{\partial y}{\partial v}  \\ \\  -\frac{\partial y}{\partial u}\end{bmatrix}, ~~~~
    g_y = \frac{1}{\widetilde{n}_z}\begin{bmatrix}-\frac{\partial x}{\partial v}  \\ \\ \frac{\partial x}{\partial u}\end{bmatrix}. \label{Eq:inverse}
\end{align}
To eliminate $f_u$ and $f_v$ from Equation~(\ref{Eq:inverse}), we derive a set of partial differential equations that are equivalent up to the choice of the coordinate system of the UV map:
\begin{align}
    \|g_x\|^2 &= \frac{1}{\widetilde{n}_x^2}\left(\left(\frac{\partial y}{\partial u}\right)^2 + \left(\frac{\partial y}{\partial v}\right)^2\right)\nonumber\\
    &= \frac{\widetilde{n}_x^2 + \widetilde{n}_z^2}{\widetilde{n}_z^2} \label{Eq:pde1}
\end{align}
because from Equation~(\ref{Eq:normal}), $\widetilde{n}_x^2 + \widetilde{n}_z^2$ can be expressed as:
\begin{align}
   \widetilde{n}_x^2 + \widetilde{n}_z^2 &= \left(\frac{\partial y}{\partial u}\frac{\partial z}{\partial v}-\frac{\partial y}{\partial v}\frac{\partial z}{\partial u}\right)^2+\left(\frac{\partial x}{\partial u}\frac{\partial y}{\partial v}-\frac{\partial x}{\partial v}\frac{\partial y}{\partial u}\right)^2\nonumber\\
    &=\left(\frac{\partial y}{\partial u}\right)^2\left(\frac{\partial z}{\partial v}\right)^2+\left(\frac{\partial y}{\partial v}\right)^2\left(\frac{\partial z}{\partial u}\right)^2-2\frac{\partial y}{\partial u}\frac{\partial y}{\partial v}\frac{\partial z}{\partial u}\frac{\partial z}{\partial v} \nonumber\\
    &+\left(\frac{\partial x}{\partial u}\right)^2\left(\frac{\partial y}{\partial v}\right)^2+\left(\frac{\partial x}{\partial v}\right)^2\left(\frac{\partial y}{\partial u}\right)^2-2\frac{\partial x}{\partial u}\frac{\partial x}{\partial v}\frac{\partial y}{\partial u}\frac{\partial y}{\partial v} \nonumber\\
    &= \left(\frac{\partial y}{\partial u}\right)^2\left(1-\left(\frac{\partial y}{\partial v}\right)^2\right) + \left(\frac{\partial y}{\partial v}\right)^2\left(1-\left(\frac{\partial y}{\partial u}\right)^2\right)\nonumber\\&-2\frac{\partial y}{\partial u}\frac{\partial y}{\partial v}\left(\frac{\partial x}{\partial u}\frac{\partial x}{\partial v}+\frac{\partial z}{\partial u}\frac{\partial z}{\partial v}\right) \nonumber\\
    &= \left(\frac{\partial y}{\partial u}\right)^2 + \left(\frac{\partial y}{\partial v}\right)^2 - 2\frac{\partial y}{\partial u}\frac{\partial y}{\partial v}\left(\frac{\partial x}{\partial u}\frac{\partial x}{\partial v}+\frac{\partial y}{\partial u}\frac{\partial y}{\partial v}+\frac{\partial z}{\partial u}\frac{\partial z}{\partial v}\right)\nonumber\\
    &= \left(\frac{\partial y}{\partial u}\right)^2 + \left(\frac{\partial y}{\partial v}\right)^2,
 \end{align}
given Equation (1), or
\begin{align}
    \|f_u\|^2 &= \left(\frac{\partial x}{\partial u}\right)^2+\left(\frac{\partial y}{\partial u}\right)^2+\left(\frac{\partial z}{\partial u}\right)^2 = 1\\
    \|f_v\|^2 &= \left(\frac{\partial x}{\partial v}\right)^2+\left(\frac{\partial y}{\partial v}\right)^2+\left(\frac{\partial z}{\partial v}\right)^2 = 1\\
    f_u^\mathsf{T}f_v &= \frac{\partial x}{\partial u}\frac{\partial x}{\partial v} + \frac{\partial y}{\partial u}\frac{\partial y}{\partial v} +\frac{\partial z}{\partial u}\frac{\partial z}{\partial v}  = 0. 
\end{align}
Similarly, we have
\begin{align}
     \|g_y\|^2 &= \frac{1}{\widetilde{n}_x^2}\left(\left(\frac{\partial x}{\partial u}\right)^2 + \left(\frac{\partial x}{\partial v}\right)^2\right)\nonumber\\
    &= \frac{\widetilde{n}_y^2 + \widetilde{n}_z^2}{\widetilde{n}_z^2}. \label{Eq:pde2}
 \end{align}
Further, the angle between the vector can be represented as:
\begin{align}
    g_x^{\mathsf{T}} g_y &= -\frac{1}{\widetilde{n}_z^2}
    \left(\frac{\partial x}{\partial v}\frac{\partial y}{\partial v}+\frac{\partial x}{\partial u}\frac{\partial y}{\partial u}\right)\nonumber\\
    &=\frac{\widetilde{n}_x\widetilde{n}_y}{\widetilde{n}_z^2} \label{Eq:pde3}
\end{align}
because $\widetilde{n}_x\widetilde{n}_y$ can be written as:
 \begin{align}
     \widetilde{n}_x\widetilde{n}_y &= \left(\frac{\partial y}{\partial u}\frac{\partial z}{\partial v}-\frac{\partial y}{\partial v}\frac{\partial z}{\partial u}\right)\left(\frac{\partial z}{\partial u}\frac{\partial x}{\partial v}-\frac{\partial z}{\partial v}\frac{\partial x}{\partial u}\right)\nonumber\\
     &=-\frac{\partial y}{\partial u}\frac{\partial x}{\partial u}\left(\frac{\partial z}{\partial v}\right)^2 - \frac{\partial y}{\partial v}\frac{\partial x}{\partial v}\left(\frac{\partial z}{\partial u}\right)^2\nonumber\\
     &+\left(\frac{\partial z}{\partial u}\frac{\partial z}{\partial v}\right)\left(\frac{\partial y}{\partial u}\frac{\partial x}{\partial v}+\frac{\partial y}{\partial v}\frac{\partial x}{\partial u}\right)\nonumber\\
     &=-\frac{\partial y}{\partial u}\frac{\partial x}{\partial u}+\frac{\partial y}{\partial u}\frac{\partial x}{\partial u}\left(\frac{\partial x}{\partial v}\right)^2+\frac{\partial y}{\partial u}\frac{\partial x}{\partial u}\left(\frac{\partial y}{\partial v}\right)^2\nonumber\\
     &-\frac{\partial y}{\partial v}\frac{\partial x}{\partial v}+\frac{\partial y}{\partial v}\frac{\partial x}{\partial v}\left(\frac{\partial x}{\partial u}\right)^2+\frac{\partial y}{\partial v}\frac{\partial x}{\partial v}\left(\frac{\partial y}{\partial u}\right)^2\nonumber\\
     &-\left(\frac{\partial x}{\partial u}\frac{\partial x}{\partial v}+\frac{\partial y}{\partial u}\frac{\partial y}{\partial v}\right)\left(\frac{\partial y}{\partial u}\frac{\partial x}{\partial v}+\frac{\partial y}{\partial v}\frac{\partial x}{\partial u}\right)\nonumber\\
     &=-\left(\frac{\partial y}{\partial u}\frac{\partial x}{\partial u}+\frac{\partial y}{\partial v}\frac{\partial x}{\partial v}\right).
 \end{align}

Equation~(\ref{Eq:pde1}), (\ref{Eq:pde2}), (\ref{Eq:pde3}) form a set of partial differential equations of $g$ with respect to the surface normal where we solve these equations by minimizing the loss of $\mathcal{L}_{\rm geo}$.

\end{document}